%% file: iccp23_template.tex
\pdfoutput=1

\documentclass[10pt,journal,compsoc]{IEEEtran}
\newif\ifpeerreview

\peerreviewfalse

\usepackage{multirow}
\usepackage[nocompress]{cite}
\usepackage{url}
\usepackage{amsmath,amssymb,graphicx}
\usepackage{ulem}
\usepackage{booktabs}
\usepackage{xcolor}
\usepackage[ruled]{algorithm2e} 
\usepackage{lipsum}
\usepackage[switch]{lineno}
\usepackage{comment}
\usepackage{xspace}
\makeatletter
\DeclareRobustCommand\onedot{\futurelet\@let@token\@onedot}
\def\@onedot{\ifx\@let@token.\else.\null\fi\xspace}

\def\eg{\emph{e.g}\onedot} 
\def\ie{\emph{i.e}\onedot}

\def\etal{\emph{et al}\onedot}
\makeatother

\input{sections/notations}

\newcommand{\paperID}{10}

\title{Omni-Line-of-Sight Imaging \\ for {\HSRfull}}

\author{
Binbin Huang$^\dagger$, ~Xingyue Peng$^\dagger$, ~Siyuan Shen, ~Suan Xia, ~Ruiqian Li, ~Yanhua Yu, ~Yuehan Wang, \\ ~Shenghua Gao, ~Wenzheng Chen, ~Shiying Li, ~and Jingyi Yu
\IEEEcompsocitemizethanks{
\IEEEcompsocthanksitem Binbin Huang and Xingyue Peng contribute equally to this work; 
\IEEEcompsocthanksitem Binbin Huang, Xingyue Peng, Siyuan Shen, Suan Xia, Ruiqian Li, Yanhua Yu, Yuehan Wang, Shenghua Gao, Shiying Li and Jingyi Yu are with School of Information Science and Technology, ShanghaiTech University;
\IEEEcompsocthanksitem Wenzheng Chen is with University of Toronto;
}
}

\begin{document}

\input{sections/0_abs}

\maketitle

\input{sections/1_introduction}
\input{sections/2_related}
\input{sections/3_problemsetup}
\input{sections/3_method}
\input{sections/4_evaluation}
\input{sections/5_real}
\ifpeerreview \else
\section*{Acknowledgments}
{\small
The authors would like to thank Linlin Li for his help with the figure, Zi Wang for helpful comments on the text, and Zhengqin Pan for his support in the data capture and initial discussion. }
\fi
\bibliographystyle{IEEEtran}
\bibliography{references}

\ifpeerreview \else






\fi

\end{document}

%% file: sections/notations.tex
\newcommand{\highlight}[1]{\textbf{#1}}

\newcommand{\ourmethod}{Omni-LOS}
\newcommand{\ourmodel}{\ourmethod}
\newcommand{\hsr}{HSR}
\newcommand{\HSRfull}{Holistic Shape Reconstruction}
\newcommand{\hsrfull}{holistic shape reconstruction}
\newcommand{\Hsrfull}{Holistic shape reconstruction}

\newcommand{\losfull}{line-of-sight}
\newcommand{\nlosfull}{non-line-of-sight}

\newcommand{\levelsetword}{level set}

\newcommand{\network}{\mathcal{F}}
\newcommand{\surface}{\mathcal{S}}

\newcommand{\point}{\mathbf{x}}
\newcommand{\direction}{\textbf{d}}

\newcommand{\levelset}{\varphi}
\newcommand{\albedo}{\rho}
\newcommand{\density}{\sigma}

\newcommand{\normal}{\mathbf{n}}
\newcommand{\occupancy}{\mathbf{o}}

\newcommand{\meas}{\tau}
\newcommand{\losmeas}{\tau_{LOS}}
\newcommand{\nlosmeas}{\tau_{NLOS}}

\newcommand{\lightfalloff}{G}
\newcommand{\lighttransport}{H}
\newcommand{\transmittance}{T}

\newcommand{\timeofflight}{t}
\newcommand{\measurepoint}{\tilde{\mathbf{x}}}
\newcommand{\spadposition}{\measurepoint_{s}}
\newcommand{\wallposition}{\measurepoint_{w}}

\newcommand{\loss}{\mathcal{L}}
\newcommand{\lsnet}{\mathcal{F}_{l}}
\newcommand{\albedonet}{\mathcal{F}_{a}}

\newcommand{\eqcomma}{\ \ ,}
\newcommand{\eqstop}{\ \ .}

\renewcommand{\paragraph}[1]{{\noindent\textbf{#1}}}

\def\lim{\mathop{\mathsf{lim}}} 



%% file: sections/0_abs.tex
\IEEEtitleabstractindextext{%
\begin{abstract}
We introduce {\ourmethod}, a neural computational imaging method for conducting {\hsrfull} (\hsr) of complex objects utilizing a Single-Photon Avalanche Diode (SPAD)-based time-of-flight sensor. As illustrated in Fig.~\ref{fig:teaser}, our method enables new capabilities to reconstruct near-$360^\circ$ surrounding geometry of an object from a single scan spot. In such a scenario, traditional {\losfull} (LOS) imaging methods only see the front part of the object and typically fail to recover the occluded back regions. 
Inspired by recent advances of {\nlosfull} (NLOS) imaging techniques which have demonstrated great power to reconstruct occluded objects, {\ourmodel} marries LOS and NLOS together, leveraging their complementary advantages to jointly recover the holistic shape of the object from a single scan position. The core of our method is to 
put the object nearby diffuse walls 
and augment the LOS scan in the front view with the NLOS scans from the surrounding walls, which serve as virtual ``mirrors'' to trap lights toward the object. 
Instead of separately recovering the LOS and NLOS signals, we adopt an implicit neural network to represent the object, analogous to NeRF~\cite{mildenhall2020nerf} and NeTF~\cite{shen2021non}. While transients are measured along straight rays in LOS but over the spherical wavefronts in NLOS, we derive differentiable ray propagation models to simultaneously model both types of transient measurements so that the 
NLOS reconstruction also takes into account the direct LOS measurements and vice versa.
We further develop a proof-of-concept {\ourmodel} hardware prototype for real-world validation. 
Comprehensive experiments on 
various wall settings demonstrate that {\ourmethod} 
successfully resolves shape ambiguities caused by occlusions, 
achieves high-fidelity 3D scan quality, and manages to recover objects of various scales and complexity. 
\end{abstract}

\begin{IEEEkeywords} 
	NLOS Imaging, Neural Rendering, Surface Reconstruction
\end{IEEEkeywords}
}

%% file: sections/1_introduction.tex
\IEEEraisesectionheading{ \section{Introduction}\label{sec:introduction}
}

\IEEEPARstart{T}{hree} dimensional (3D) shape reconstruction in computer vision has largely followed ``to reconstruct is to see'' strategy, \ie, only parts that are observable from the camera can be recovered. 
\emph{\Hsrfull} ({\hsr}) aims to recover the $360^\circ$ surrounding geometry of an object.
As a long-standing problem, it plays a key role in various applications, 
ranging from shape scanning in virtual and augmented reality to archaeological preservation and even rescue missions~\cite{cipresso2018past,wojciechowski2004building,daly2004digital}.

Traditionally, {\hsr} adopts the implicit assumption that the object can be observed from front to back and up to bottom, \ie, the object is scanned in an omni-directional way.
The capture can be conducted by either using a hand-held camera~\cite{cui20103d} or constructing specialized imaging apparatus such as $360^\circ$ camera arrays~\cite{joo2015panoptic,kim2009multi} or a mechanically rotating board~\cite{NextEngine}, \cite{matterandform}.
However, obtaining omni-directional imagery data towards the target object in reality can be challenging and sometimes prohibitive.
For example, a hand-held camera requires a user to hold a camera and circle around an object, which takes tremendous user effort. 
Alternatively, specialized imaging apparatuses are expensive and have sophisticated synchronization or calibration procedures. 

\begin{figure*}[t]
	\centering
	\includegraphics[width=\textwidth]{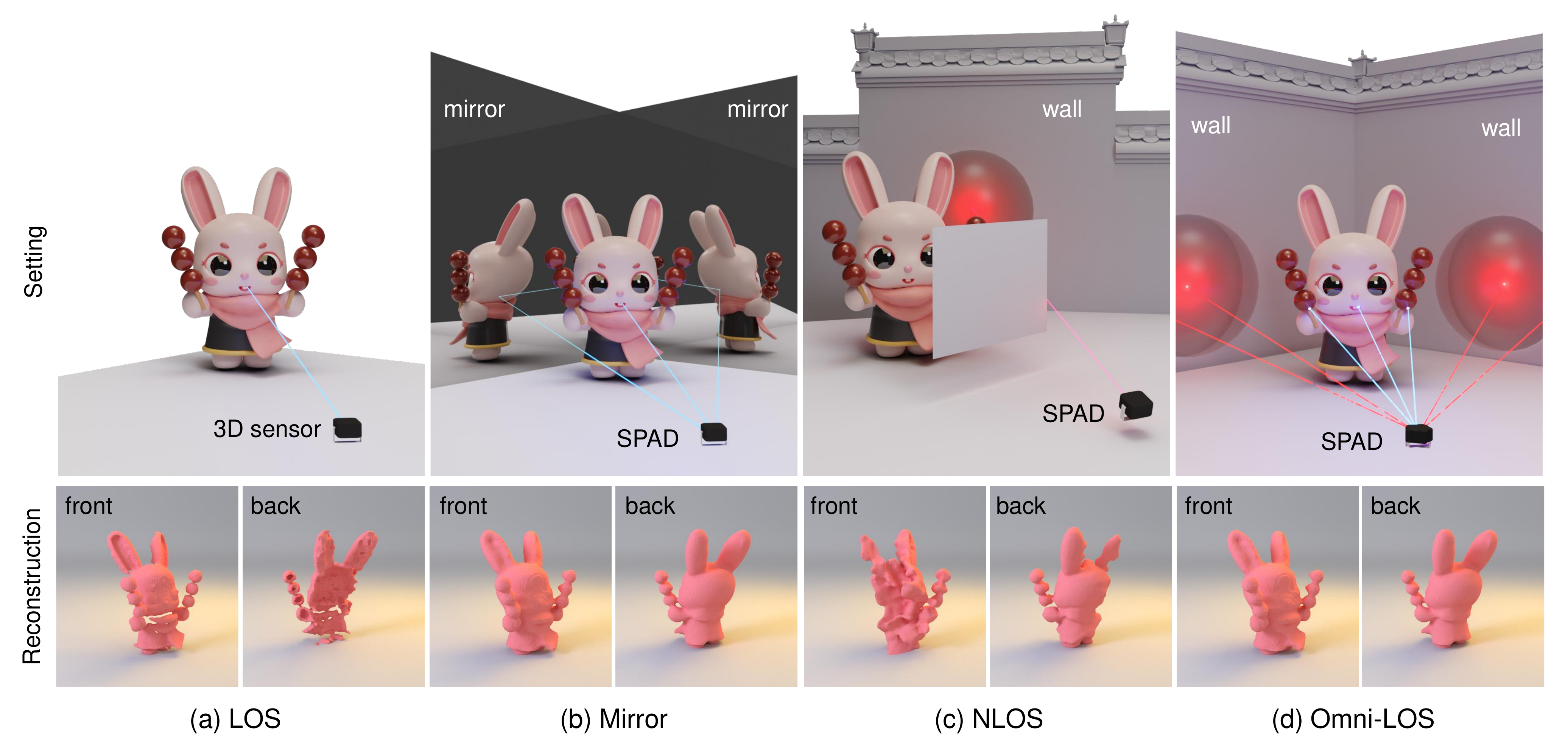}
 \vspace*{-1cm}
	\caption{\textbf{Conceptual illustration of 4 different  setups under single-scan {\hsr} task}. When we observe an object from a single scan spot, 
    (a) LOS imaging techniques only see the front shape but fail to recover the back part. (b)  Mirror systems (\eg~\cite{Xu2018CVPR}) help create virtual back views to reconstruct the holistic shape but suffer from calibration and correspondence matching issues. (c) NLOS imaging methods treat diffuse walls as ``diffuse mirrors'' to reconstruct the occluded object. Still, they only recover the regions that are visible to the wall.  
    (d) Our {\ourmethod} system enables new capabilities to conduct HSR from a single scan spot, where we apply LOS signals to recover the front, visible regions, as well as NLOS measurements to reconstruct the occluded, back part. 
 }
	\label{fig:teaser}
 \vspace*{-0.5cm}
\end{figure*}

There have been a number of attempts to overcome the omni-directional scanning requirement and explore ways to conduct HSR from a single scan spot. 
In such a scenario, a sensor watches an object at a fixed position only (Fig.~\ref{fig:teaser}).
While simplifying the capture condition, due to the limited field of view, only the front part of the object appears in the direct line-of-sight (LOS) view of the sensor. 
As such, HSR becomes challenging since it lacks the necessary information to reconstruct the occluded back geometry.
Existing methods propose to use auxiliary mirrors to reflect lights toward the object,  so that the back part beyond LOS regions can be virtually observed~\cite{Miraldo2022CVPR,Takahashi2022PAMI,Xu2018CVPR}. 
Such approaches are applicable to both regular RGB cameras~\cite{Miraldo2022CVPR, Takahashi2022PAMI} as well as the time-of-flight (ToF) cameras~\cite{ Xu2018CVPR}. 
Albeit effective, these methods inherently require mounting mirrors, and for large-scale objects, the approach can be too expensive in both cost and labor.

We recognize the single-scan HSR problem partially resembles the non-line-of-sight (NLOS) imaging setting,
where both aim to reconstruct an object that is occluded and not directly observable from the senor. 
In the single-scan HSR case, the occlusion comes from the object itself, \eg, the back geometry is occluded by the front part. 
Instead, in the NLOS imaging setting, it is typically due to an occluder.
Effective NLOS solutions are largely attributed to the latest advances in imaging hardware and a range of inverse imaging techniques. 
Particularly, by exploiting the laser and single-photon avalanche diode (SPAD) to prob the scene, as well as developing powerful inversion techniques such as LCT~\cite{2018LCT}, $f$-$k$~\cite{2019FK} or Phasor Field~\cite{liu2020phasor, 2019Liu}, NLOS techniques have demonstrated excellent capabilities to reconstruct occluded objects faithfully.

In theory, existing NLOS techniques can be deployed to holistic shape recovery. By placing an object nearby diffuse walls (Fig.~\ref{fig:teaser}), one can
employ LOS methods to reconstruct the front, visible part while applying NLOS methods to recover the back, occluded geometry. In reality, naively combining two reconstruction results incurs multiple problems: 
1) Due to the inconsistency of separate reconstruction results, seams might happen across views, especially in the boundaries between LOS and NLOS scans. 
2) Part of the NLOS signal comes from the front, visible regions of the object, which also appear in the LOS measurements but are not effectively used in NLOS reconstruction and vice versa; 
3) LOS and NLOS measurements have distinct ray propagation models but there is no unified framework to simultaneously handle both of them. 

In this paper, we present {\ourmethod}, a {\hsrfull} technique that combines LOS and NLOS imaging under a unified neural modeling framework.
As shown in Fig.~\ref{fig:teaser}, {\ourmethod} employs a SPAD and a laser emitter to prob an object.
It fixes the sensor in front of the object and effectively employs the surrounding walls as virtual mirrors, eliminating the need to move the sensors or use tailored mirror systems.
At the core of our method is a neural representation~\cite{mildenhall2020nerf,shen2021non} that encodes the holistic shape of the object with {\levelsetword} methods~\cite{wang2021neus}. Following the ``analysis-by-synthesis'' principle, we derive differentiable ray propagation models for both LOS and NLOS transient imaging procedures and integrate them into the training pipeline, supervising the network parameters by minimizing the differences between the rendered transient images and the real measurements.
By simultaneous training on both LOS and NLOS signals, {\ourmethod} enables accurate and reliable holistic geometry and normal recovery, even when observing the object from a single scan spot.

We conduct comprehensive experiments to validate {\ourmethod} on object reconstruction under different wall settings, including a single wall, two side walls, and with and without a ceiling.
Compared with the brute-force techniques that individually conduct reconstruction on LOS and NLOS scans and then fuse the results, we show {\ourmethod} significantly improves the quality of the recovered shape, especially on boundary regions across different scans.
We further construct proof-of-concept hardware prototypes to demonstrate our approach in real captures.
Experiments on synthetic and real scenes demonstrate  {\ourmethod} achieves significantly better 3D scan quality 
and may provide a practical omni-directional reconstruction solution suitable for recovering objects of various scales.  

The main contributions of this work can be summarized as follows.

\begin{itemize}
\item We introduce {\ourmethod}, an end-to-end solution that enables HSR from a single scan spot. It largely simplifies the capture condition, requiring neither camera arrays nor surrounding mirror systems.

\item  {\ourmethod} derives differentiable ray propagation models for both LOS and NLOS transient imaging procedures, as well as proposes an implicit level-set representation to recover shape jointly from  LOS and NLOS scans. Compared to brute-force methods, our unified framework demonstrates significantly better reconstruction quality.

\item We further build a real-world imaging system and validate {\ourmethod} with real captures, demonstrating its potential usage for real-world deployment. The code and data will be publicly available to the community to stimulate future developments in HSR reconstruction.

\end{itemize}

%% file: sections/2_related.tex
\section{Related Work}
\label{sec:related}
\paragraph{{\HSRfull} (\hsr).}
{\hsr} relies on trapping light of the $360^\circ$ geometry of an object, \ie, observing it in an omni-directional way. In traditional line-of-sight (LOS) settings, due to the limited field of view, typically multi-view images are captured to cover the full body of the object. The images can be either taken from hand-held cameras~\cite{cui20103d,barron2022mipnerf360,pollefeys2004visual}, or special imaging apparatuses like $360^\circ$ camera arrays~\cite{joo2015panoptic,kim2009multi,zhou2022relightable,cai2022humman}. 
With the acquired multi-view images, 
techniques such as multi-view stereo (MVS) methods~\cite{hartley2003multiple,pollefeys2004visual,schonberger2016structure} or neural modeling approaches~\cite{mildenhall2020nerf,mueller2022instant,barron2022mipnerf360} can be applied to infer the underlying geometry from appearance. However,  obtaining omni-directional imagery data sometimes is challenging in reality, potentially due to the tremendous user efforts or the expensive hardware cost. 
To overcome the onmi-directional scanning requirement, people propose to watch an object from a single spot but utilizing auxiliary sources to serve as ``virtual cameras'', which can also provide the onmi-directional information of the object. 
Two typical examples are mirrors and walls, which we discuss below. 

\vspace*{0.3cm}
\paragraph{Trapping Light with Mirrors.}
Intuitively, due to the special reflective property, a mirror can be treated as a virtual camera. The idea of using mirrors to synthesize more views and applying them in 3D reconstruction has been proposed for a long time, 
ranging from regular RGB cameras~\cite{miraldo2022unified,Ying2010CVPR,Gluckman2002PAMI,fang2021reconstructing} to time-of-fight (ToF) cameras~\cite{Xu2018CVPR}. 
Most existing works~\cite{miraldo2022unified,Ying2010CVPR,Gluckman2002PAMI,fang2021reconstructing} focus on extending more field of view with mirrors. Similar to us, 
Xu~\etal~\cite{ Xu2018CVPR} augment a ToF sensor with a multi-mirror system to conduct HSR of an object from a single observation spot, where in the mirror system, the light rays would bounce multiple times within the trap volume to eventually reach nearly all regions on the object, especially the side and the back part. 
However, mirror-based methods typically have challenging calibration and correspondence matching problems and are not suitable for performing reconstruction applications in the wild.

\vspace*{0.3cm}
 \paragraph{Trapping Light with Walls.}
 Instead of real mirrors, people also explore using the relay walls as ``diffuse mirrors'', which diffusely scatter light from the object. 
The scattered signals can be used to recover the object nearby the wall.
Such works are often explored in non-line-of-sight (NLOS) imaging, where the goal differs slightly from {\hsr}. NLOS imaging aims to reconstruct an occluded object and not directly observable by the sensor. 
Due to the weakness of the collected signals, time-resolved imaging systems are often employed in NLOS tasks, such as streak cameras \cite{Kirmani2009ICCV,Velten2012NC, 2014Wu, 2021Feng}, photonic mixer devices \cite{Heide2013ToG, OToole2014ToG}, and SPADs \cite{2015Mauro, 2019FK, 2018LCT, 2019Ahn, 2019Xin, 2019Liu, shen2021non, 2021Pei, 2021Peng, 2021Nam}. Our method follows this fashion and employs a laser emitter and a SPAD to capture transient information of an object.

Previous NLOS methods represent the scene by either volumetric albedo~\cite{2018LCT,2019FK}, volumetric feature~\cite{Mu2022PAMI,chen2020ToG} or parametric surface~\cite{2019Xin,Tsai2019CVPR}. Recently, neural-based approaches have made remarkable progress~\cite{shen2021non, fujimura2023nlosneus}. 
NeTF~\cite{shen2021non} 
applies a density field to represent the scene, which fails to acquire a high-quality surface. Concurrent to our work, NLOS-NeuS~\cite{fujimura2023nlosneus} represent NLOS scenes with signed distance fields (SDF)~\cite{wang2021neus, Vicini2022sdf}, which shows better surface recovery. 
Different from them, our technique incorporates a more general {\levelsetword} representation, on which we develop a unified transient rendering formulation under both LOS and NLOS settings. Our method not only achieves high-quality surface reconstruction but also eliminates the common Eikonal regularization term and stabilizes the network training.

\vspace*{0.3cm}
\paragraph{Joint Reconstruction with LOS and NLOS Scans.}
There have been several pioneering works that try to combine LOS and NLOS imaging together. 
Henley~\etal~\cite{Henley2022TCI} use a SPAD and a laser to estimate the shape of a 3D scene from the first- and second-bounce measurements, where the hidden objects in the cast shadows can be recovered from the second-bounce measurements. Rapp~\etal~\cite{Rapp2020NC} propose an edge-resolved transient imaging (ERTI) method to reconstruct the visible and hidden parts of the scene, where they scan the scene along an arc 
around a vertical edge and analyze the transient histogram differences to reconstruct the hidden objects. 
We differ from them in both problem setup and methodology. {\ourmethod} tackles the single-scan {\hsr} problem. Moreover, we develop a neural modeling technique to jointly utilize both LOS and NLOS transients to reconstruct the holistic geometry of the object. 

%% file: sections/3_problemsetup.tex
\section{{\ourmethod}   Framework}
We consider the problem of single-scan \hsr, where the goal is to reconstruct the whole shape of an object observed from \emph{a single scan spot}.
As shown in Fig.~\ref{fig:teaser}, it is extremely challenging for traditional LOS imaging methods (Fig.~\ref{fig:teaser}a). Due to the limited fields of view, LOS methods can only recover the front, visible shape.
Prior works propose to augment new views with mirrors~\cite{Xu2018CVPR} (Fig.~\ref{fig:teaser}b). Such methods successfully estimate the complete geometry but suffer from calibration and correspondence-matching issues.  
Inspired by NLOS imaging methods (Fig.~\ref{fig:teaser}c) which estimate hidden objects with three-bounce transient information from diffuse walls,
we propose {\ourmethod}~ (Fig.~\ref{fig:teaser}d), a brandy new framework to conduct {\hsr}~
by utilizing cues from both LOS and NLOS measurements. 

As illustrated in Fig.~\ref{fig:teaser}d, we place an object nearby diffuse walls and employ a single-photon avalanche diode (SPAD) together with a laser emitter to shoot rays and capture transient images of the object. 
{\ourmethod} applies LOS methods to estimate the visible, red regions, as well as taking the nearby walls as “diffuse mirrors” and employing NLOS techniques to infer the occluded, blue regions.  
We mainly consider a two-wall setting, which we find enough to cover the near $360^\circ$ geometry of the object. However, it can be easily applied to various wall settings like single-wall or with-ceiling cases. We validate different wall settings in the experimental sections (Sec.~\ref{sec:syntheticexp} and Sec.~\ref{sec:real}). 

To better utilize cues in both LOS and NLOS imaging measurements, 
we propose a neural modeling framework to represent the object with {\levelsetword} based implicit neural networks. 
We further derive differentiable ray propagation models for both LOS and NLOS imaging conditions and jointly train the model from all measurements. 
We find that joint training achieves a significant performance boost compared to the separate reconstruction baseline. 

In the below sections, we first detail our {\levelsetword}-based neural representation and the differentiable ray propagation models in Sec.~\ref{sec:method}. 
Next we analyze our method on a synthetic dataset in Sec.~\ref{sec:syntheticexp}. 
Finally, we build a proof-of-concept {\ourmethod} hardware prototype and verify its effectiveness with real-world experiments in Sec.~\ref{sec:real}.

\begin{figure*}
	\centering
	\includegraphics[width=\textwidth]{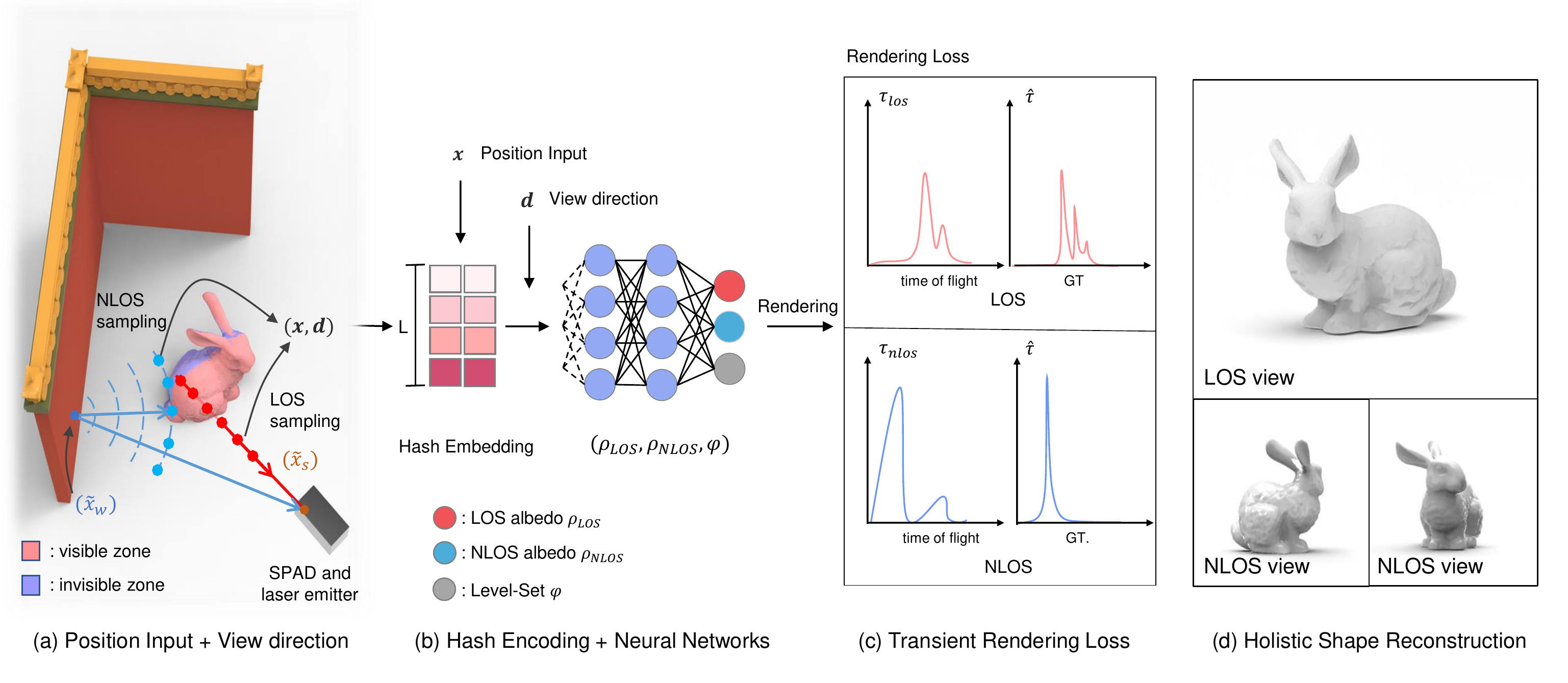}
    \vspace*{-1cm}
    \caption{
    \textbf{{\ourmethod} Pipeline} 
    (a) We first sample 3D points along straight rays in LOS measurements (red, visible regions) while over spherical wavefronts in NLOS measurements (blue, occluded regions). 
    (b) These 3D samples are then fed into a hash encoding network~\cite{mueller2022instant} to predict the corresponding level set values and albedo. To account for the light intensity difference between LOS and NLOS, we predict a 2-channel albedo to represent them separately. 
    (c) We render the samples under both LOS and NLOS imaging settings and compare them with the real measurements. 
    (d) By representing the object with a unified neural network and optimizing from both LOS and NLOS signals, {\ourmethod} successfully reconstructs the high-quality, holistic shape of the object. 
 }
 \vspace*{-0.3cm}
	\label{fig:pipe}
\end{figure*}

%% file: sections/3_method.tex
\section{\ourmethod ~Shape Recovery}
\label{sec:method}
\subsection{Neural {\levelsetword} Representation}
\label{sec:neuralrepre}
Neural representations have emerged as a powerful tool to represent objects or scenes with complex appearance, geometry and material~\cite{mildenhall2020nerf,wang2021neus,2021physg}. 
Following this fashion, we also employ neural representations as a flexible and compact parameterization method to encode the holistic shape of an object.  
Specifically, we choose a {\levelsetword} based representation for better geometry modeling~\cite{yariv2020multiview,wang2021neus}:
\begin{equation}
    \network_{\Theta}: (\point, \direction) \rightarrow (\levelset(\point), \albedo(\point, \direction)) \eqcomma
\end{equation}
where we employ a neural network $\network_{\Theta}$ which takes as input position $\point=(x,y,z)$ and ray direction $\direction=(d_x, d_y, d_z)$ and predicts both the level set value $\levelset(\point)$ and the view-dependent surface albedo $\albedo(\point, \direction)$. 
Here, the {\levelsetword} value $\levelset(\point)$ implicitly defines the surface at zero, \ie, the surface $\surface = \{ \point | \levelset(\point) = 0 \}$. 
The {\levelsetword} values in the outside regions are defined as positive while the inside values are negative. 
Following IDR~\cite{yariv2020multiview}, the surface normal ${\normal}$ can be obtained by 
\begin{equation}
    \normal(\point) = \frac{\nabla_{\point}\levelset(\point)}{\|\nabla_{\point}\levelset(\point)\|} \eqstop
\end{equation}

The {\levelsetword} representation also enables flexible conversion to an occupancy indicator function~\cite{occ19lars}, where the inside region is one while the free space is zero. We adopt a sigmoid function to transform {\levelsetword} values to the indication of occupancy:
\begin{equation}
    \occupancy(\levelset) = \frac{1} {1+e^{s\levelset}} \eqcomma
    \label{eq:occupancy}
\end{equation}
where $s$ is a scalar 
that controls the sharpness of the sigmoid function.  

We highlight that both normal $\normal$ and occupancy $\occupancy$ are crucial in our transient ray propagation models, which are described in Sec. {\ref{sec:holistic_transient_ rendering}} and Sec.~\ref{sec:transmattance}, respectively. First, normal $\normal$  participates in calculating the radiometric quantities. 
Moreover, occupancy $\occupancy$ plays an important role in density modeling in the ray propagation equation(Eq.~\eqref{eq:rendering}). Incorporating them in the neural {\levelsetword} representations 
allows for reliable shape and accurate surface reconstruction compared to the density volume representation~\cite{mildenhall2020nerf}.

In contrast to NeuS or NLOS-NeuS that directly use a signed distance field, the level set is more flexible as it doesn't require the magnitude of the gradient to be constant, eliminating the requirement of tuning the common Eikonal regularization term. We provide a detailed analysis in the supplementary. 

\vspace*{-5pt}
\subsection{Neural Transient Rendering}
\label{sec:holistic_transient_ rendering}
We now describe how to render the level set representation into transient measurements under both LOS and NLOS imaging conditions. Different from regular RGB images, transient images can be described as $ \meas(\measurepoint, \timeofflight)$, where $\measurepoint$ is the light source and camera positions (\ie, in LOS case, it denotes the SPAD position $\spadposition$ while in NLOS case it is typically referred to a wall position $\wallposition$ (See Fig.~\ref{fig:pipe}a), $\timeofflight$ is the light traveling time. Here, we consider a confocal setting~\cite{2018LCT} in both LOS and NLOS imaging, where the SPAD sensor and the camera are located at the same position. Note that our framework can be generalized to the non-confocal setting as well. 

Rendering neural representations into transient images has been explored in NeTF~\cite{shen2021non}, where Shen~\etal choose a density-based neural representation and focus on NLOS setting only. 
Instead, we derive both LOS and NLOS neural transient rendering equations under level set representations. 
Our formula demonstrates better geometry and normal reconstruction in NLOS imaging over NeTF, as well as enables HSR from a single scan spot when LOS data is applicable.

 Without loss of generality, we define LOS and NLOS image formation models in a unified integral equation: 
\begin{equation}
\begin{aligned}
	\meas(\measurepoint, \timeofflight) = 
 \underset{\Omega} 
 {\int} \lighttransport(\measurepoint, \point) 
 \lightfalloff( \measurepoint, \point)\transmittance^2( \measurepoint, \point) \density(\point)\albedo(\point, \direction)
 d\point  \eqstop
\end{aligned}
\label{eq:rendering}
\end{equation}

$\Omega$ is the 3D sampling space.  In LOS setting it is a straight ray while in NLOS setting it is a spherical wavefront. $\lighttransport(\measurepoint, \point)$ is Dirac term of the light transport, \eg, the time that light travels from light source $\measurepoint$ to the 3D point $\point$ and back to the sensor position $\measurepoint$ should be equal to $\timeofflight$. 
$\lightfalloff(\measurepoint, \point)$ describes the radiometric fall-off due to the light traveling and surface reflections. 
$\sigma(\point), \rho(\point, \direction)$ is the volumetric density 
and view-dependent albedo. 
$\transmittance(\measurepoint, \point)$ is a transmittance term representing the mutual visibility from the sensor to the 3D surface point: 
\begin{align}
\transmittance(\measurepoint, \point) = \mathrm{exp}{\left(-\int_{0}^{\frac{\timeofflight}{2}}\density(\spadposition+sc\direction) ds \right)} \eqstop
\label{eq:transparency}
\end{align}
We apply a square $\transmittance^2(\measurepoint, \point)$ in the integral to describe the active imaging process, where light travels from the SPAD to the 3D point and then reflects back to the sensor. 
$c$ is the constant light speed.
In the following, we separately define the rendering equations for both LOS and NLOS settings. We include detailed derivation in the supplementary. 

\vspace*{0.2cm}
\paragraph{LOS Transient Rendering Equation.} 
The LOS rendering equation can be described as:
\begin{equation}
\begin{aligned}
	\losmeas(\spadposition, \timeofflight) =  \int_{\Omega}\delta(\| \spadposition - \point \| - \frac{1}{2}\timeofflight c) 
 \\
 \frac{(\normal\cdot \widehat{\point\spadposition})
 }{\|\point-\spadposition\|^2}
 \transmittance^2(\spadposition, \point)
 \density(\point)\albedo(\point, \direction)d\point  \eqcomma
 \end{aligned}
	\label{eq:rendering:los}
\end{equation} 
where we measure the transient intensity of a ray shooting from spad position  $\spadposition$ along direction $\direction=\widehat{\point\spadposition}$ at time $\timeofflight$, \ie, $\point=\spadposition + \frac{1}{2}tc\direction$. Note that here we fix the SPAD position $\spadposition$ but varying view direction $\direction$ to capture different 3D points. 
The Dirac term $\lighttransport=\delta(\left | \spadposition - \point \right | - \frac{tc}{2} )$ constrains the light travel time from $\spadposition$ to the 3D point $\point$ should be $\frac{t}{2}$. 
The light fall-off $\lightfalloff(\point, \spadposition) = {(\normal\cdot \widehat{\point\spadposition})}/{\|\point-\spadposition\|^2}$. This formulation resembles LOS rendering in T\"oRF~\cite{attal2021torf}. 
The key difference is that we further add normal in the light intensity calculation. 
Besides the distance, we consider the cosine foreshortening term $(\normal\cdot \widehat{\point\spadposition})$ and find it brings more accurate surface reconstruction. Implement-wise, we ensure $(\normal\cdot \widehat{\point\spadposition})$ to be larger than zero by truncating the negative part. 

\vspace*{0.2cm}
\paragraph{NLOS Transient Rendering Equation.}
On the other side,  the NLOS rendering equation is defined as:
\begin{multline}
	\nlosmeas(\wallposition, t) = \underset{\Omega}{\int}  \delta(\| \wallposition - \point  \|-\frac{ct}{2}) \cdot \\ 
\frac{(\normal\cdot\widehat{\point\wallposition})^2(\normal_{w}\cdot\widehat{\wallposition\point})^2
}{\|\point-\wallposition\|^4}
 \transmittance^2(\wallposition, \point) \density(\point) \albedo(\point, \direction)  d\point\eqstop
	\label{eq:rendering:nlos}
\end{multline}
Here, we measure the transient intensity of a ray scatters from a relay wall point $\wallposition$ to 3D space and reflect back to   $\wallposition$ again at a specific time step along $t$. 
The Dirac term $\lighttransport= \delta(\| \wallposition - \point  \|-\frac{ct}{2})$. 
After scattering three times, the light fall-off is 
${(\normal\cdot\widehat{\point\wallposition})^2(\normal_{w}\cdot\widehat{\wallposition\point})^2
}/{\|\point-\wallposition\|^4}
$ where $\normal_{w}$ denotes the surface normal at the relay wall spot $\wallposition$. 
As we consider at most three bounces, the light travels as a spherical wavefront in NLOS. Following NeTF~\cite{shen2021non}, we uniformly sample in $(\theta, \phi)$ space for computing the spherical integral. 
Compared to~\cite{shen2021non}, we incorporate surface normal in transient rendering, which enables more accurate surface reconstruction. 
Note that the view-dependent model allows {\ourmethod} to represent various materials. We demonstrate it in Sec.~\ref{sec:real} 

\vspace*{0.0cm}
\subsection{Volumetric Rendering of Neural Surface }
\label{sec:transmattance}
We now discuss how to model density  $\density$ in the rendering equations. 
In contrast to \cite{attal2021torf,shen2021non} that aims to reconstruct volumetric fields, 
we aim to reconstruct an implicit surface parameterized by neural networks. To do so, 
we should model the volumetric density $\density(\point)$ as a function of an implicit surface $\levelset(\point)$. 
In the following, we re-parameterize all the 3D points $\point$ along a ray as $\point(\eta)=\measurepoint + \direction \eta$. We slightly abuse the notations $\transmittance(\eta) = \transmittance(\measurepoint, \point(\eta))$ and density  $\density(\eta) = \density(\point(\eta))$ 
for clarity. 
We define the weight term $w(\eta)$ of $\point(\eta)$ as 
\begin{equation}
    w(\eta) = \transmittance^2(\eta)\sigma(\point(\eta))\eqcomma
    \label{eq:weight}
\end{equation}
where $w(\eta)$ weights radiometric intensity to the transient measurements. Since we aim to reconstruct the surface, we should ensure the radiometric contribution mostly comes from the surface $\surface$, \ie, the zero level set. Following~\cite{wang2021neus}, we first assume there is only a single visible plane when ray marches from outside into inside. 
With this assumption, one can define the $w(\eta) = \delta(\levelset(\point))$ so that, ideally, only points at the zero-level set contributes to transient measurement. To apply differentiable rendering, we seek a smooth Dirac approximation $\delta_s$ instead of $\delta$. We notice that the predefined occupancy indicator $\occupancy$ (Eq.~\eqref{eq:occupancy}) meets our purpose  
as it is a smooth Heaviside step function and its derivative serves as a smooth Dirac function $\delta_s$. Therefore, we model $w(\eta)$ as a function of {\levelsetword} $\levelset$ by
\begin{equation}
    w(\eta) = \delta_s(\levelset(\point(\eta))) = -\frac{d \occupancy(\levelset(\point(\eta)))}{d\eta} 
    \label{eq:transformation}
\end{equation}
By substituting Eq.~\eqref{eq:occupancy} and Eq.~\eqref{eq:weight} into Eq.~\eqref{eq:transformation} and solving the PDE, we can derive the density $\density$ as 
\begin{align}
 \density(\point(\eta)) & = 
 \frac{-s}{2(1+e^{-s\levelset(\point)})}\nabla_\point\levelset(\point)\cdot \direction \eqstop
 \label{eq:density}
\end{align}
Notice that the density is only valid for a single plane intersection where the ray travels from outside to inside. To account for multi-plane intersection, we  adopt $\hat{\density} = \max(\density, 0)$~\cite{wang2021neus}  
and composite transmittance with Eq.~\eqref{eq:transparency} with $\hat{\density}$. 

Remarkably, our method differs from existing neural surface rendering
~\cite{wang2021neus,yariv2021volume}. While they employ signed distance fields that assume $\|\nabla \levelset(\point)\|=1$ everywhere, our formulation is based on a more general level set function and has incorporated $\nabla\levelset(\point)$ in our forward model, so we do not need explicitly impose eikonal regularization~\cite{gropp2020implicit}. 
See the supplementary ablation study of {\levelsetword} representation. 

\vspace*{0.2cm}
\paragraph{Weight Warm-up}. 
We notice that the weight of a sample in Eq.~\ref{eq:transformation} is essentially view-dependent, \ie, only locations where $(-\normal \cdot \direction) > 0$ have positive weights. If the network is not well initialized, there may be many points not being optimized at all. To ensure every point stands a chance for optimization, we apply a warm-up trick. We define 
\begin{equation}
    w_\text{train}(\eta) = \alpha w(\eta) + (1-\alpha)\|\nabla\occupancy(\eta)\|\eqcomma
\end{equation}
where $\|\nabla\occupancy(\eta)\|$ is a view-independent density. During training, we first ignore the influence of the normal to get a good level set field initialization. We increase $\alpha$  from $0$ to $1$ to switch from the view-independent model to the view-dependent model. 

\begin{figure}
    \centering
    \includegraphics[width=\columnwidth]{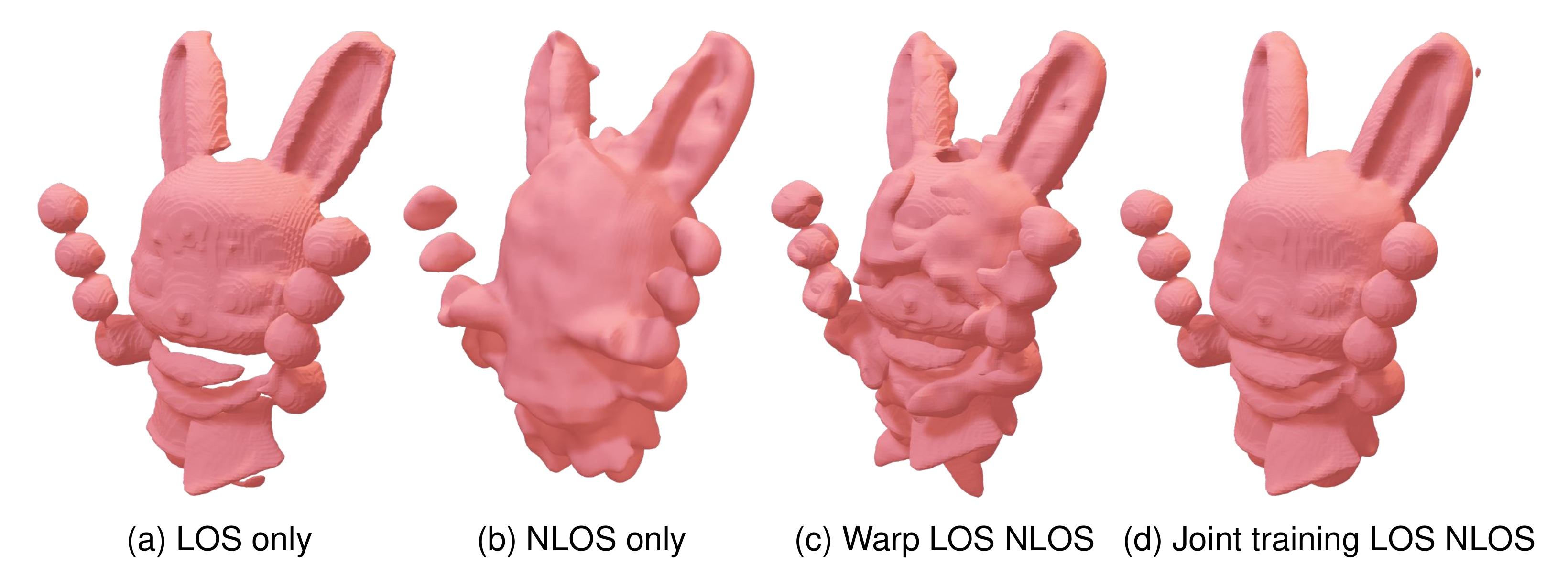}
    \vspace*{-1cm}
    \caption{\textbf{Joint Training}. {\ourmethod} supports joint training with both LOS and NLOS measurements and demonstrates significant improvements over the naive baseline that separately reconstructs LOS and NLOS partial shapes and fuses them together. 
    (a,b) Training with only LOS or NLOS data results in good front or back shape reconstruction. However, the other side is missing. 
    (c) Fusing the LOS and NLOS shape together has severe merging artifacts, especially for the edge regions between two scans. 
    (d) Our joint training framework demonstrates much more accurate shape reconstruction. 
    }
    \label{fig:joint}
    \vspace*{-0.3cm}
\end{figure}

\subsection{Learning Objectives}
Our method allows joint training of both LOS and NLOS transient images. 
However, the measurement intensity in NLOS and LOS images are quite different, \ie, NLOS intensity is much weaker. 
Therefore, 
in each 3D point, we predict a 2-channel surface albedo $\rho$ to account for LOS and NLOS intensity separately. Note that the geometry representation $\levelset(\point)$ is shared to ensure holistic surface reconstruction from multiple sources.  
\par

Besides, we find LOS measurements provide strong geometric priors. Encouraged by \cite{tsai2017geometry}, we detect the first returning photon from the LOS transient and create a 3D binary mask $M$ to indicate occupancy. 
That's to say, for the regions where LOS data didn't detect any object, those regions should be empty ($\occupancy=0$). We supervise the network by minimizing the binary cross entropy (BCE) loss between $M$ and the predicted occupancy value. Adding mask priors brings faster convergence speed and better shape recovery. 
Taking all these into consideration, 
we set our learning objectives as:
\begin{equation}
\small
\begin{aligned}
	\loss_{b} & =\sum_{\wallposition}\| \meas_{NLOS}(\wallposition, t) - \bar{\meas}_{NLOS}(\wallposition, t)\|^2 \\ & + \gamma \sum_{\spadposition}\|\meas_{LOS}(\spadposition, t) - \bar{\meas}_{LOS}(\spadposition, t)\|^2 \\
 & + \beta \sum_{\spadposition,t}\text{BCE}(\occupancy(\levelset(\point_{\eta})), M(\spadposition,\eta
))
 \eqcomma
\label{eq:loss}
\end{aligned}
\end{equation}
where $\gamma$ and $\beta$ are to balance the objectives and $\point_\eta$ a sample point along the ray and $M$ is the corresponding binary mask. We empirically set $\gamma = 1$ and $\beta = 0.01$. Fig.~\ref{fig:joint} demonstrates the superiority of our joint training framework.

\subsection{Network Implementation}
We design our network based on the fast neural training framework i-NGP~\cite{mueller2022instant}. 
It consists of a learnable hash table and two shallow MLPs representing geometry $\lsnet$ and directional albedo $\albedonet$. We set the hash table with $L=32$ and $D=8$ to output a $256$-dim hash embedding. 
$\lsnet$ is a three-layer MLP that takes as input the hash embedding and outputs a 1-dim level set value and a 15-dim geometry feature. $\albedonet$ is a two-layer MLP that takes the geometry feature and the encoded view direction and outputs a 2-channel surface albedo (representing both LOS and NLOS) with a Softplus activation. 
A detailed network architecture is included in the supplementary materials.

For each iteration, we sample a location from SPAD positions $\spadposition$ or wall positions $\wallposition$ to conduct LOS or NLOS training. Specifically, for LOS training, we sample  $256$ points along a ray to render $\meas_{LOS}(\spadposition, t)$. 
For NLOS training, we sample $64\times64\times128$ points, where  $64\times64$ is the $\theta, \phi$ angles of the wavefront and $128$ is different radius samples (for transmittance computation). 
We adopt Adam \cite{kingma2014adam} as our optimizer with a learning rate beginning at $1 \times 10^{-3}$ and decaying exponentially to $5 \times 10^{-5}$. By default, we conduct the training with 100k iterations and find that 10k iterations can produce a reliable result (roughly 10 mins) for most experiments. After optimizing the networks, we infer a $256^3$ level set grid and extract a mesh using marching cubes. All experiments are implemented based on Pytorch and run on a server with an NVIDIA A40 GPU. 

%% file: sections/4_evaluation.tex
\begin{figure*}[t]
    \centering
    \includegraphics[width=1.0\textwidth]{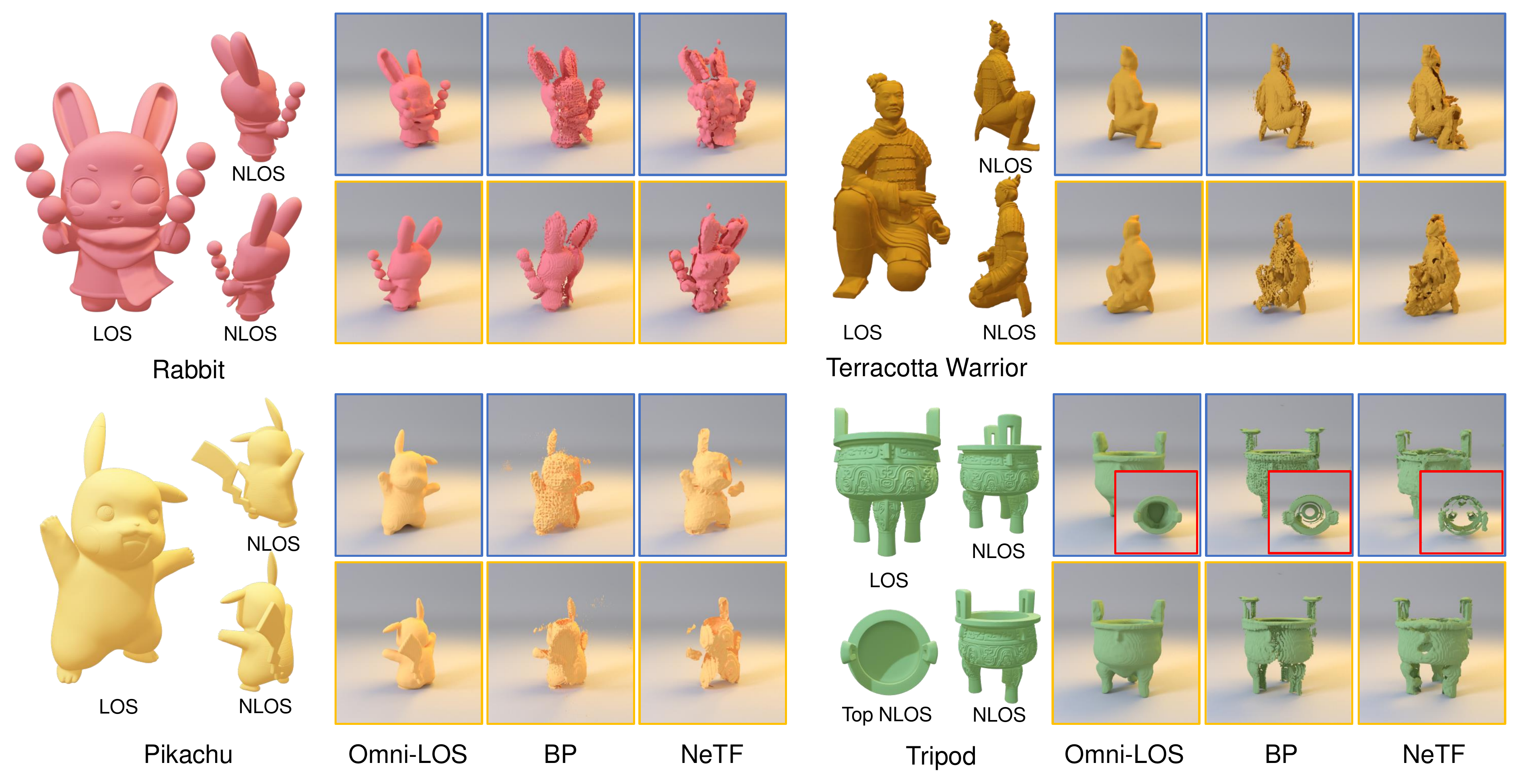}
    \vspace*{-1cm}
    \caption{ \textbf{Synthetic Evaluation}. 
    We show 4 reconstruction examples. For each example, left is its capture setting. On the right we show the reconstructed shape in both front and back views. Our method is able to reconstruct high-quality, continuous and smooth surfaces, while the baselines produce poor bumpy surfaces and have artifacts in the edges between LOS and NLOS scans. We apply two diffuse walls for the rabbit, the terracotta warrior, and the Pikachu examples. For the tripod, we additionally add a ceiling as another NLOS relay wall to recover the concave shape. 
    }
    \label{fig:gallery}
\end{figure*}

\vspace*{-2pt}
\section{Evaluation}
\label{sec:syntheticexp}
\subsection{Sythetic Datasets}
We first evaluate our method under synthetic data to quantitatively analyze the performance of our method. As we tackle a new single-scan HSR setting, we need to synthesize LOS measurements in the front view and NLOS measurements in multiple back views (assuming two walls). 
We adopt a publicly available NLOS transient renderer from ~\cite{chen2020ToG} and extend it to synthesize both multi-view NLOS transients and LOS transients. 

We first normalize the object and put it at the center of a $1m \times 1m$ ground. We then build two $1m \times 1m$ walls at the edges of the ground. 
From each planar wall, we acquire transients at a resolution of $256 \times 256 \times 512$. 
To simulate LOS transients, we sample along a ray instead of a wavefront 
and uniformly synthesize $256\times256$ transients. 

In the real world, the object might occlude the walls. However, many NLOS techniques, \eg, LCT or $f$-$k$~\cite{2018LCT,2020DLCT}, require regular sampling on the relay wall to reconstruct the hidden object. 
To compare with those existing NLOS methods, we consider an ideal scenario. We assume that the walls are fully visible and synthesize transients at all spots on the relay walls. We also omit the inter-reflections between different walls in NLOS measurements. Note our technique also supports training with irregular samples on the wall. In Sec.~\ref{sec:real}, we demonstrate the capability of our method under a real setting. 

\renewcommand{\tabcolsep}{3pt}
\begin{table}[t]
\caption{\textbf{Quantitative Comparisons}. We measure the Chamfer Distance (cm) and Normal Consistency (\%) between the predicted shape and GT shape. {\ourmethod} outperforms other methods for most cases.}
  \centering
  \resizebox{.95\columnwidth}{!}{
  \begin{tabular}{llcclcclcclcc}
    \toprule
     \multirow{3}{*}{Methods} &&\multicolumn{2}{c}{Rabbit}&&\multicolumn{2}{c}{Pikachu} &&\multicolumn{2}{c}{Terracotta Warrior} &&\multicolumn{2}{c}{Tripod}   \\
     \cmidrule(){3-4} \cmidrule(){6-7} \cmidrule(){9-10} \cmidrule(){12-13}   
   &&CD$\downarrow$ & NC $\uparrow$ && CD$\downarrow$& NC $\uparrow$ && CD$\downarrow$& NC $\uparrow$  && CD$\downarrow$& NC $\uparrow$ \\
    \cmidrule(){1-1} \cmidrule(){3-13}
    DLCT~\cite{2020DLCT} && 2.28 & 77.59 && 1.73 & 85.38 && 1.46 &77.45 && \highlight{1.63}& 79.61\\
    $f$-$k$~\cite{2019FK} &&  1.94 & 67.73 && 1.32& 72.81  && 1.11 &69.43 && 1.68& 70.68\\
    PF~\cite{liu2020phasor} && 2.07 &63.83 && 1.62 &67.92  && 1.43 &61.81 && 2.76& 59.59 \\
    BP~\cite{Velten2012NC} && \highlight{1.79}&71.64 && 1.34 & 74.64 && 1.30 &71.61 && 1.92& 71.79\\
    NeTF~\cite{shen2021non} && 2.10& 69.01 && 1.86 &71.07  &&  1.88 &71.67 && 2.56& 72.65\\
    Ours && \highlight{1.79}& \highlight{85.93} && \highlight{0.09}&\highlight{94.01} && \highlight{1.02} &\highlight{87.69}  && 1.87&  \highlight{83.02}\\
    \bottomrule
    \end{tabular}
    }
    \label{tab:quantitative}
\end{table}

\begin{figure}[t]
    \centering
\includegraphics[width=1.0\columnwidth]{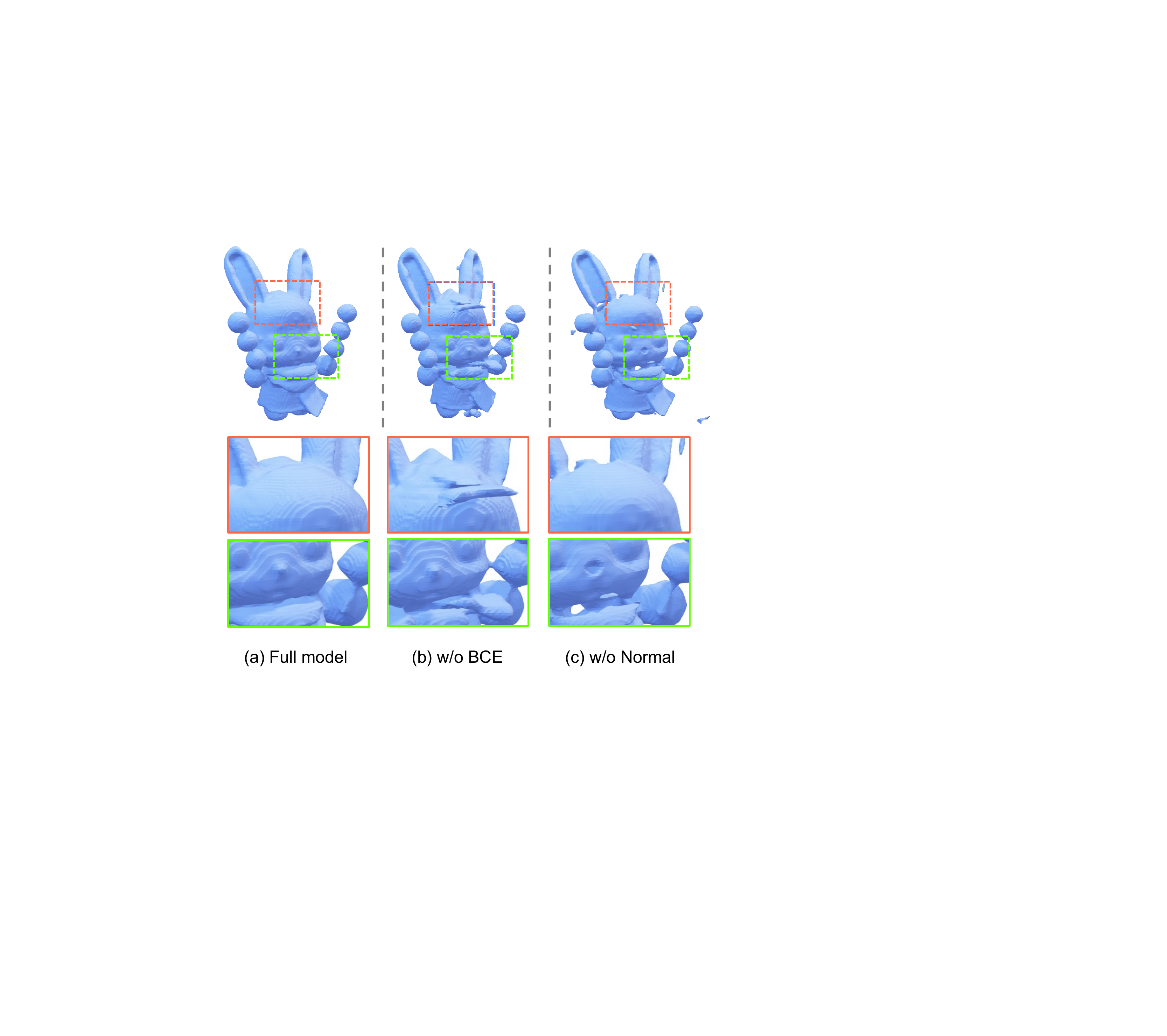}
 \vspace*{-0.8cm}
    \caption{\textbf{Ablation Study for BCE loss and Normal}. (a) Our full model effectively recovers the accurate surface of the object. (b) When the BCE loss is removed, some floating regions appear. (c) Without the normal constraint in the image formation models, the recovered surface becomes concave, such as the nose and the neck of the rabbit.}
     \vspace*{-0.5cm}
    \label{fig:albation}
\end{figure}

\subsection{Omni-LOS Surface Reconstruction}
Our method applies {\levelsetword} implicit representations, which can be easily converted into explicit meshes using marching cubes. 
We demonstrate our method acquires more accurate surface reconstruction over prior arts. 
We mainly consider NeTF~\cite{shen2021non} as the neural-based baseline as it also applies neural representation.
To make a fair comparison, we train NeTF with the same network architecture as us but change the prediction to be density values and apply its own image formation model. Since NeTF is designed for NLOS only, we apply NeTF to reconstruct each wall separately, then extract multiple surfaces and the LOS surface and fuse them together. We do the same things for DLCT~\cite{2020DLCT} since it also produces NLOS surfaces.

We further compare our technique to other volume-based NLOS methods, including Back-projection (BP)~\cite{Velten2012NC}, $f$-$k$~\cite{2019FK}, Phasor Field (PF)~\cite{liu2020phasor}, and LCT~\cite{2018LCT}. 
We apply them to reconstruct the NLOS volume from each wall separately. For LOS transients, 
we warp it into a new volume with Back-projection. 
Given all volumes, we fuse them into a single volume by summation. We then apply marching cubes to the fused volume to acquire the holistic shape of the object. Due to limited space, we put some qualitative results in the supplementary. In the main paper, we only show the comparison with BP and NeTF. 

\begin{figure}[t]
    \centering
    \includegraphics[width=\columnwidth]{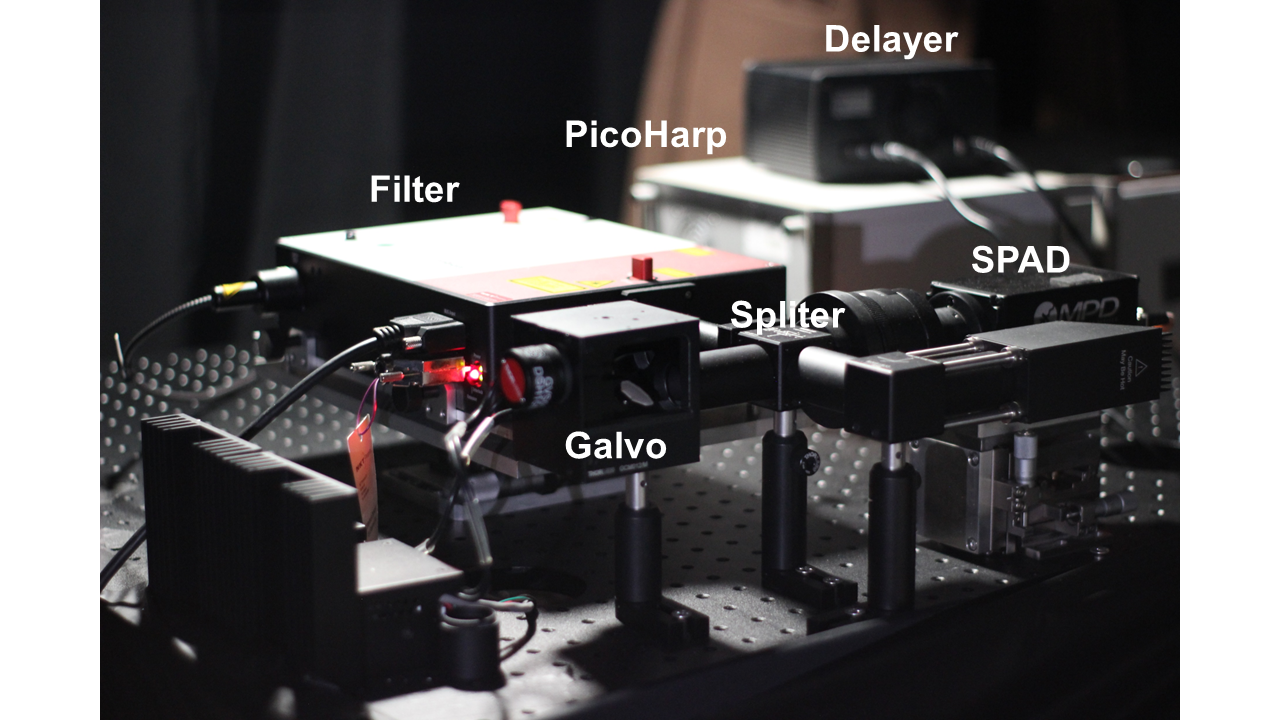}
    \caption{\textbf{Capture System}. Our prototype includes a pulsed laser, a SPAD with TCSPC for photon counting, and a galvo for scanning on the relay wall. }
    \label{fig:device}
\end{figure}

We show qualitative results in  Fig.~\ref{fig:gallery}. We also quantitatively evaluate the shape reconstruction quality in Tab.~\ref{tab:quantitative}. 
For quantitative comparison, we check the chamfer distance (CD)~\cite{qi2016pointnet} and the normal consistency (NC)~\cite{groueix2018} between the predicted shape and GT shape. 
We uniformly sample 20K points 
from the reconstruction mesh and ground truth shape, which we denote as $\point\in\mathcal{X}$ and $\mathbf{y}\in\mathcal{Y}$. Chamfer distance (CD) and normal consistency (NC) are defined as 
\begin{equation}
\footnotesize
\begin{aligned}
    \text{CD}(\mathcal{X}, \mathcal{Y}) &= \frac{1}{2}\text{D}(\mathcal{X}, \mathcal{Y}) + \frac{1}{2}\text{D}(\mathcal{Y}, \mathcal{X}) \\
    \text{D}(\mathcal{X}, \mathcal{Y}) &= \frac{1}{\|\mathcal{X}\|}\sum_{\point\in\mathcal{X}}\|\point-q_{\mathcal{Y}}(\point)\| \eqstop
 \end{aligned}
\end{equation}%
\begin{equation}
\footnotesize
\begin{aligned}
    \text{NC}(\mathcal{X}, \mathcal{Y}) &= \frac{1}{2}\text{N}(\mathcal{X}, \mathcal{Y}) + \frac{1}{2}\text{N}(\mathcal{Y}, \mathcal{X}) \\
    \text{N}(\mathcal{X}, \mathcal{Y}) &= \frac{1}{\|\mathcal{X}\|}
    \sum_{\point\in\mathcal{X}}\|\normal(\point)\cdot\normal(q_{\mathcal{Y}}(\point))\|~\eqstop
\end{aligned}
\end{equation}
Here, $q_{\mathcal{Y}}(\point)$ means finding the closest point $\mathbf{y}\in\mathcal{Y}$ from $\point$.

Several interesting observations can be found in the results. 
First, our method achieves better reconstruction quality. As for the whole shape, {\ourmethod} achieves more accurate and complete shape recovery, even for the extremely challenging concave tripod example. In contrast, baseline methods typically have large holes or missing regions. We believe the success comes from our {\levelsetword} representation and the unified neural modeling framework. On the contrary, naively reconstructing the LOS and NLOS part separately and merging them together produces strong artifacts, especially for the edges between LOS and NLOS scans. When checking the LOS and NLOS surface reconstruction quality, our neural modeling method produces a smooth surface while baselines get bumpy or pixelate surface. 
In quantitative comparison (Tbl.~\ref{tab:quantitative}), we have better scores in Rabbit, Pikachu and Terracotta Warrior examples, while it slightly drops for the Tripod example. We hypothesize it is because all the methods fail for the bottom part of the tripod since it is not observed at all. As a result, this part strongly influences the score. Although $f$-$k$ and DLCT are better than us, the three methods are pretty close and our normal is much better. 

Lastly, we find our results lose some fine details in Terracotta Warrior or Tripod examples, potentially due to the limited measurement resolution. Further improving our technique to recover fine details is an interesting direction and we leave it for future work. 

\begin{figure}[t]
\centering
\includegraphics[width=1.0\columnwidth]{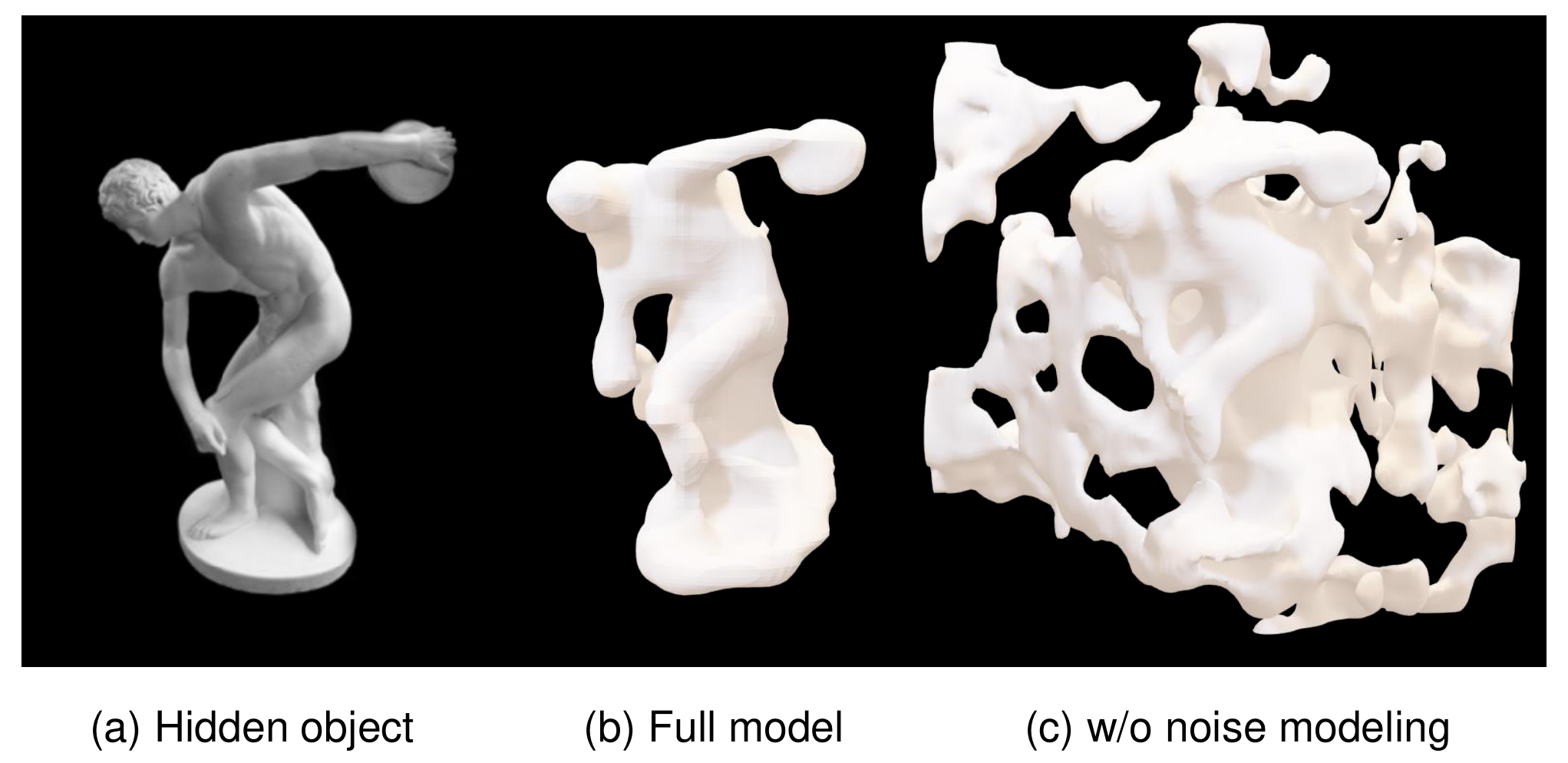}
\vspace*{-0.2cm}
    \caption{\textbf{Noise Handling}. We add a noise prediction branch to account for noisy NLOS transients. We verify it on an NLOS scene~\cite{2019FK}. Our noise modeling successfully removes most artifacts degraded by NLOS noise. }
    \label{fig:noise}
\end{figure}

\subsection{Ablations}

We apply normal in the image formation models (Eq.~\eqref{eq:rendering:los} and Eq.~\eqref{eq:rendering:nlos}) and BCE mask in the objective function (Eq.~\ref{eq:loss}). 
We ablate them two in Fig.~\ref{fig:albation}.  
The normal guides the surface directions (Eq.~\eqref{eq:density}) 
and helps recover a more accurate surface. Ignoring normal in the rendering equation may 
generate a surface with incorrect curvature, \eg, the concave shape shown in Fig.~\ref{fig:albation}. 

Conversely, BCE loss helps ``carve'' the space and gets clean reconstruction results. Since the field of view of LOS scan is quite small, removing BCE loss makes no constraints for the unobservable regions and generates floating artifacts in those regions. 

%% file: sections/5_real.tex
\section{Validation with Prototype}
\label{sec:real}

\begin{figure*}[t]
    \centering
    \includegraphics[width=0.8\textwidth]{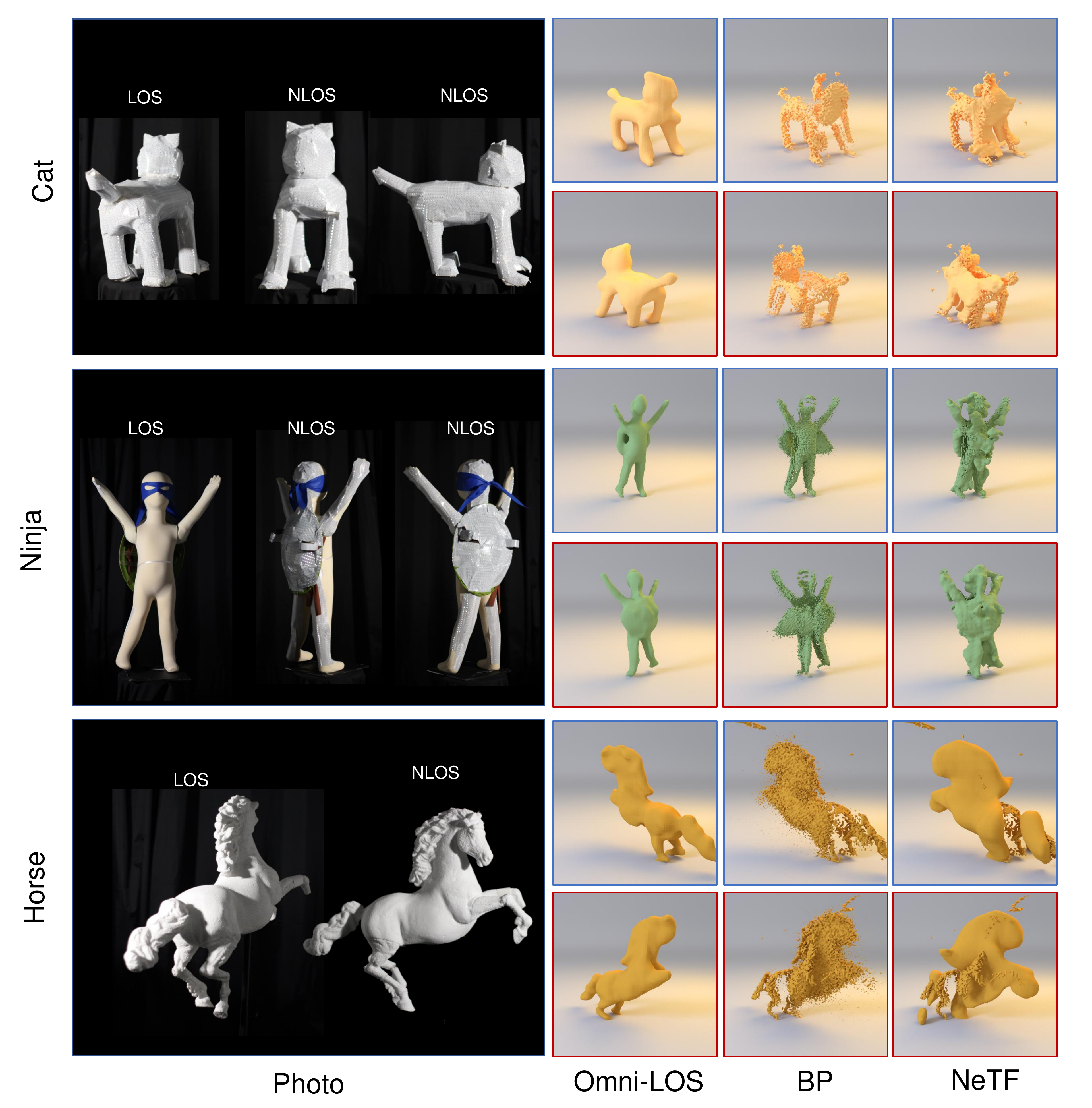}
    \caption{\textbf{Real World Experiments}. We validate  {\ourmethod} on three objects with different shapes and materials. The cat is made of retro-reflective material and has a relatively simple geometry. Our approach precisely recovers the cat's head, feet, and tail. 
    The ninja is made of cloth, while the turtle shell is covered with retro-reflective material. {\ourmethod} recover both the front of the ninja and the back of the turtle shell. 
    The horse is made of foam and therefore presented as the most challenging case due to the low SNR. Our approach recovers the head, back foot, and even the horse's tail. 
However, it fails to recover the horse's front foot due to the limited number of returning photons. %
On the other side, baseline methods generate poor geometry reconstruction with missing regions, large holes, and pixelated artifacts in all 3 cases.}
    \label{fig:real}
\end{figure*}

\subsection{Hardware}

Fig.~\ref{fig:device} shows our experimental setup. A supercontinuum laser (SuperK EXTREME FIU-15) with a wavelength-tunable filter (SuperK VARIA) emits a collimated light beam. We use the wavelength band between 665 nm and 675 nm with an average power of 48 mW and a repetition frequency of 39 MHz. The beam passes through a polarized beam-splitter (Thorlabs VA5-PBS251) and a 2D galvanometer (Thorlabs GVS012) and scans the relay wall. 
Light bounces off diffusely three times: on the relay wall, object surface, and back on the relay wall. A fast-gated SPAD coupled with a PicoHarp 300 at 4 ps resolution then collects photons that return through the galvanometer. We exploit a delayer (PicoQuant PSD-065-A-MOD) to shield the LOS transients with a gate width of 12 ns and record NLOS transients only. Our system is confocal, i.e., the laser spots coincide with the detection points on the relay walls. The hardware is about 1.5 m from the scanning area of $1 m \times 1 m$ on each wall; the objects, e.g., cat, ninja, horse, are 0.5 m away from the relay walls. 

Unlike traditional NLOS cases, our single-scan HSR setting involves measuring an object both directly and from multiple relay walls. We calibrate our system on each relay wall using the onsite method \cite{pan2022onsite} and scan the object on a uniform $128\times128$. 
grid. We then split all the measured points into irregular LOS and NLOS scans by determining whether the point hit the wall or not. 
While most NLOS methods only support regular grid sampling measurements~\cite{2018LCT,2019FK}, our technique allows irregular sampling at arbitrary points. 
When we capture under multiple walls, different walls will interfere with each other in NLOS measurements. We apply LOS measurements to roughly divide the object range and manually crop the interference NLOS signals from other walls. We do this for all the methods. We compare our method with BP and NeTF, as they also support irregular sampling measurements. 

\subsection{Handling Noise}
NLOS transients typically contain strong noise due to the weak signals and the imperfection of the hardware systems, \eg, timing jitter, pile-up or the intractable multi-bounce scattering in the NLOS scene. 
To handle the noise, following NeRF in the wild\cite{martinbrualla2020nerfw}, we employ another network branch to predict the residual error. This procedure resembles transient rendering but renders noise instead of radiance fields:
\begin{equation}
\small
\meas_\text{noise}(\measurepoint) = \underset{\Omega}{\int}\delta(\| \wallposition - \point  \|-\frac{ct}{2}) 
\pi(\point)d\point \eqcomma
\label{eq:noise}
\end{equation}
where we predict noise $\pi(x)$ for each 3D point $\point$. Then we add it to the transient to get the final prediction: $\hat{\meas_{NLOS}}(\measurepoint, t) = \meas_{NLOS}(\measurepoint, t) + \meas_\text{noise}(\measurepoint)$. We further regularize the noise intensity to be small: 
\begin{equation}
\small
\loss_\text{NLOS}= \sum_{\wallposition}\| \meas_\text{NLOS}(\wallposition, t)-\hat{\tau_\text{NLOS}}({\wallposition}, t)\| + \lambda\|\meas_\text{noise}(\measurepoint) \|^2 \eqcomma
\end{equation} where $\lambda$ is the noise coefficient related to the signal-to-noise ratio (SNR). Transients with a higher SNR should impose a larger $\lambda$ to weaken the effect of noise. 
The noise branch only exists in the training phase.
During inference, we ignore the noise branch and acquire the shape from the regular geometry and albedo branches. 
Our technique enables reliable holistic shape even from noisy transients. Note that we predict the noise only for NLOS scans since the LOS transients are clean enough. We show an ablation study in Fig.~\ref{fig:noise}, where we apply our method on a purely NLOS statue scene~\cite{2019FK}. Implementation details are included in Supplementary Material. 

\subsection{Validation} 
{\ourmethod} models the view-dependent effects and can therefore support various materials. 
We validate our method on three objects with different materials: a retro-reflective cat, a ninja carrying a turtle shell, and a foam horse. The first two examples are captured with two NLOS walls. As the horse has a slender body, we tried a different wall setting and only applied the NLOS data from the right wall. Fig.~\ref{fig:real} shows the qualitative reconstruction results. In the retro-reflective cat example, our method successfully reconstructs the complete and clear shape, including the slender tail, limbs, and ears. This demonstrates our method's ability to handle objects of different materials. BP and NeTF lose details of the object, such as the cat's ears. In the case of the ninja which contains complex geometry and material (the human is plastic while the turtle is retro-reflective), our method also successfully reconstructs the limbs and complete turtle shell,  as well as the gaps between the human body and the shell. BP fails to reconstruct the circular turtle shell while NeTF reconstructs a messy, rough geometry. In the case of the foaming horse, {\ourmethod} produces a continuous geometry. In contrast, both comparative methods show seams between the LOS and NLOS parts, as well as pixelated  LOS reconstruction results. 

\section{Conclusion}
\begin{figure}[t]
    \centering
    \includegraphics[width=0.9\columnwidth]{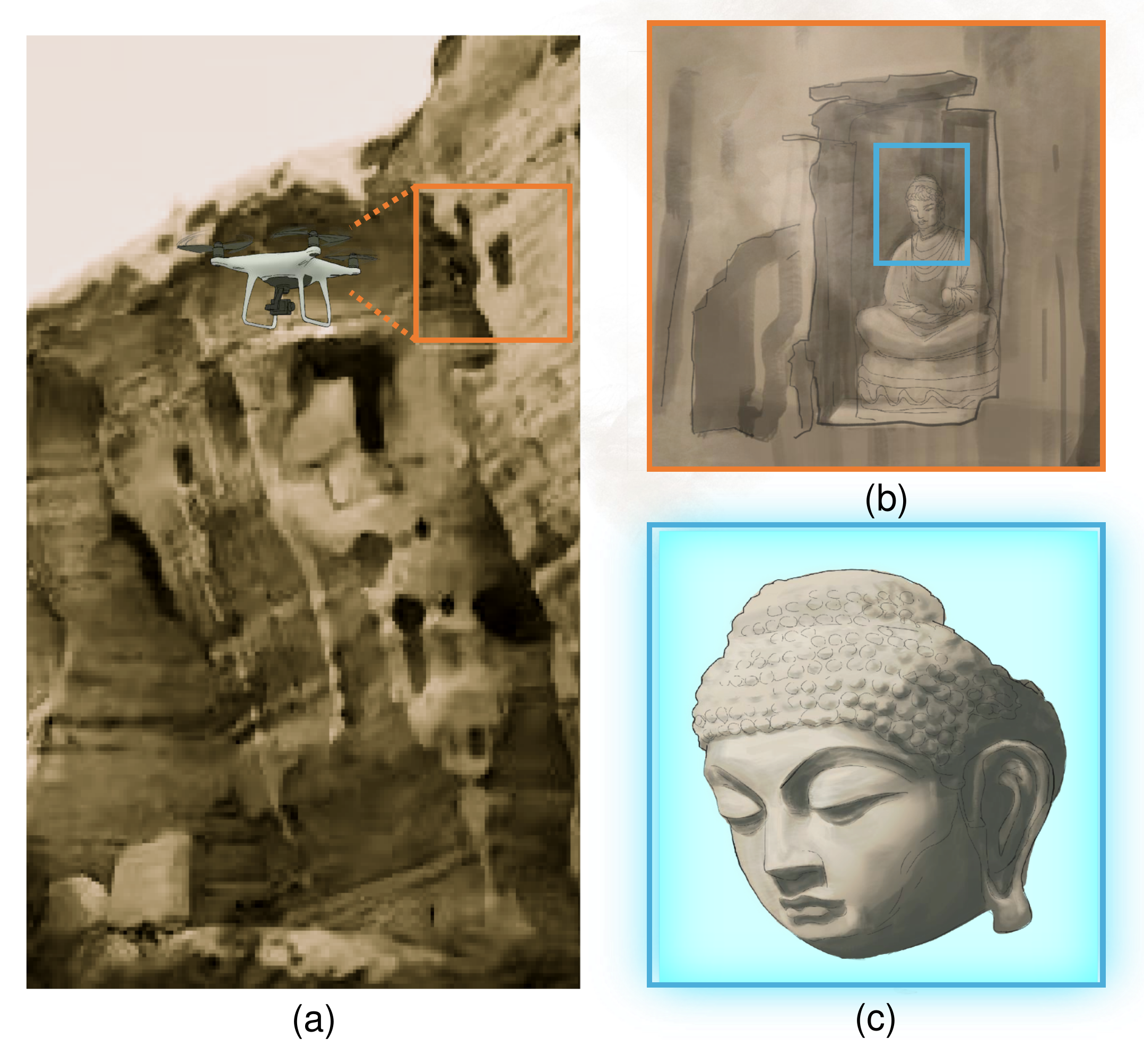}
    \caption{Our method has the potential for various future applications, especially for archaeological tasks.  
    (a) we show a flying drone probing a dangerous historic site on the cliff. (b) is the enlarged photograph of the orange region. 
    (c) common LOS-based reconstruction methods only recover the front face of the statue, while {\ourmethod}  can 
 be applied to reconstruct the whole shape of the statue, even its back part is hidden within cavernous recesses.}
    \label{fig:application}
\end{figure}
3D shape reconstruction plays an important role in computer vision, with various applications ranging from virtual and augmented reality to archaeological preservation and rescue missions. 
In this paper, we proposed {\ourmethod}, a novel approach that enables {\hsrfull} 
from a single scan spot. Our method opens a new door for in-the-wild reconstruction tasks. For example, in some challenging and dangerous cases like archaeological scenes or rescue missions, our system allows for flexible full-shape reconstruction from the front view, eliminating the requirements of archaeologists or rescue members going to dangerous, unknown back regions. Future work can focus on developing a more portable and efficient system that can be equipped in the real world, as well as further improving the reconstruction quality to recover fine details of the object. 